\documentclass[Quan,preprint, 12pt]{elsarticle}

\let\today\relax
\makeatletter
\def\ps@pprintTitle{%
	\let\@oddhead\@empty
	\let\@evenhead\@empty
	\def\@oddfoot{\footnotesize\itshape
		{Submitted preprint} \hfill\today}%
	\let\@evenfoot\@oddfoot
}
\makeatother

\usepackage{amssymb}
\usepackage{amsmath}
\usepackage{multirow}
\usepackage{graphicx}
\usepackage{float}
\usepackage{algorithm}
\usepackage[noend]{algpseudocode}
\usepackage{subcaption}

\begin{document}

\begin{frontmatter}



\title{Optimised Least Squares Approach for Accurate Polygon and Ellipse Fitting}

\author[label1,label2]{Yiming Quan\corref{cor1}}
\ead{yiming.quan@foxmail.com}
\tnotetext[cor1]{corresponding author.}
\author[label2]{Shian Chen}
\affiliation[label1]{organization={Department of Civil Engineering, Engineering College, Lishui University},
	addressline={No 1 Xueyuan Road}, 
	city={Lishui},
	postcode={323000}, 
	state={Zhejiang},
	country={China}}
\affiliation[label2]{organization={Ningbo Tianyi Design Research of Surveying and Mapping Co. Ltd},
	addressline={2F Huashang Mansion, No 100 Xinghai South Road}, 
	city={Ningbo},
	postcode={315040}, 
	state={Zhejiang},
	country={China}}

\clearpage

\begin{abstract}
This study presents a generalised least squares based method for fitting polygons and ellipses to data points. The method is based on a trigonometric fitness function that approximates a unit shape accurately, making it applicable to various geometric shapes with minimal fitting parameters. Furthermore, the proposed method does not require any constraints and can handle incomplete data. The method is validated on synthetic and real-world data sets and compared with the existing methods in the literature for polygon and ellipse fitting. The test results show that the method achieves high accuracy and outperforms the referenced methods in terms of root-mean-square error, especially for noise-free data. The proposed method is a powerful tool for shape fitting in computer vision and geometry processing applications.
\end{abstract}




\end{frontmatter}

\section{Introduction}
\label{Intro}
Fitting observed points to geometric shapes is a common problem in science and engineering fields. Among various shapes, circle and ellipse fitting has received much attention from researchers \cite{Mai, Abdul-Rahman, Prasad, Wang, Maalek}, and these shape fitting methods have been applied in diverse fields, such as robotics \cite{Dong}, engineering measurement \cite{Quan, Lu}, and medicine \cite{LiJ, Ranjbarzadeh}. Polygon fitting, notably rectangles, is equally important for applications in fields such as computer vision \cite{Yang, Liu, Bazin}, medicine \cite{Stroppa}, and remote sensing \cite{Seo, Feng}, where polygonal shapes represent common artifacts. Hence, it is essential to develop generalised and accurate methods for fitting polygons and ellipses to observed data points.

Many studies have attempted to address the problem of fitting polygonal shapes. Some of these methods applied genetic algorithms (GA) to approximate the polygons \cite{Stroppa, Yin, Yin2, Tsai, LiuWatson}. Although GA based methods are effective in solving multi-modal optimization problems, the implementation of GA requires more computation resources and is time-consuming. Liparulo et al. \cite{Liparulo} proposed to use fuzzy algorithm to reduce the computational cost of point-to-polygon distance estimation, but the main focus of the method was on shape recognition rather than shape fitting. Werman and Keren \cite{Werman} developed a Bayesian paradigm for parametric and non-parametric fitting of rectangle to noisy data points. Minimum bounding rectangle (MBR), proposed in \cite{Freeman} and \cite{Chaudhuri2007}, has been extensively used for object recognition from LiDAR point clouds \cite{Feng, Kwak2012, Kwak2014, Kabolizade}, images \cite{Yang, WangY, Xia, Ding, Liu}, and videos \cite{Edgcomb, Hu}. However, MBR aims to enclose all the points rather than match their distribution. 

Least squares methods are widely used for shape fitting. Prasad et al. \cite {Prasad} proposed a fast least squares based ellipse fitting method that does not require constraints and iterations. Chaudhuri et al. \cite{Chaudhuri2011} introduced a rectangle fit method with a bisection of upper and under-estimated rectangles. The iteration scheme of Chaudhuri's method is based on the computation of rectangle area.  However, Yang and Jiang \cite{Yang} suggested that this method could be improved for discrete and noisy data points, because the area-based fitting metric might not capture the true shape of the data. Another approach to fit rectangle, proposed by Sampath and Shan \cite{Sampath} was to fit line segments and group them by slopes with parallelism, and then determine the bounding by least squares with perpendicularity constraints. Building on Chaudhuri's and Sampath and Shan's work, Seo et al. \cite{Seo} introduced a rectangle model with eight parameters (angle and distance for each edge) and used least squares adjustment to fit the model to points, subject to the constraint of perpendicularity between edges.  Similarly, Sinnreich \cite{Sinnreich} employed least squares, but reduced the number of parameters to five, corresponding to the degrees of freedom of a symmetrical polygon, and derived a simpler design matrix based on continuous hypotrochoid functions. Stroppa et al. \cite{Stroppa} regarded Sinnreich's algorithm as the most effective method for fitting polygons. However, Stroppa et al. \cite{Stroppa} reported that hypotrochoid functions caused errors at the vertices of polygons.

Motivated by these observations, this study introduces a generalised least squares based shape fitting method that seeks to overcome the limitations of the existing methods. The objectives of this paper are to: i) present a generalised fitting algorithm that can be applied to polygons and ellipses; ii) improve the least squares based polygonal fitting algorithm with a more accurate continuous fitness function; iii) test the accuracy of the proposed method in fitting shapes to clean and noisy data. The rest of this paper is organised as follows. Section \ref{FitnessFunction} presents the generalised fitness function for polygonal and elliptical shapes. Section \ref{EstPhi} describes the calculation of the angle parameter in the fitness function. Section \ref{lsadjust} describes the least squares shape fitting algorithm based on the fitness function. Section \ref{implement} explains the implementation of the algorithm. Section \ref{validation} describes the simulated and real tests, and reports the testing results and analyses. Section \ref{conclusions} concludes the main outcome of this study.

\section{Least squares polygon and ellipse fitting}
\subsection{A generalised fitness function for unit shapes}
\label{FitnessFunction}
\begin{figure}[h!]
	\centering
	\includegraphics[width=1\linewidth]{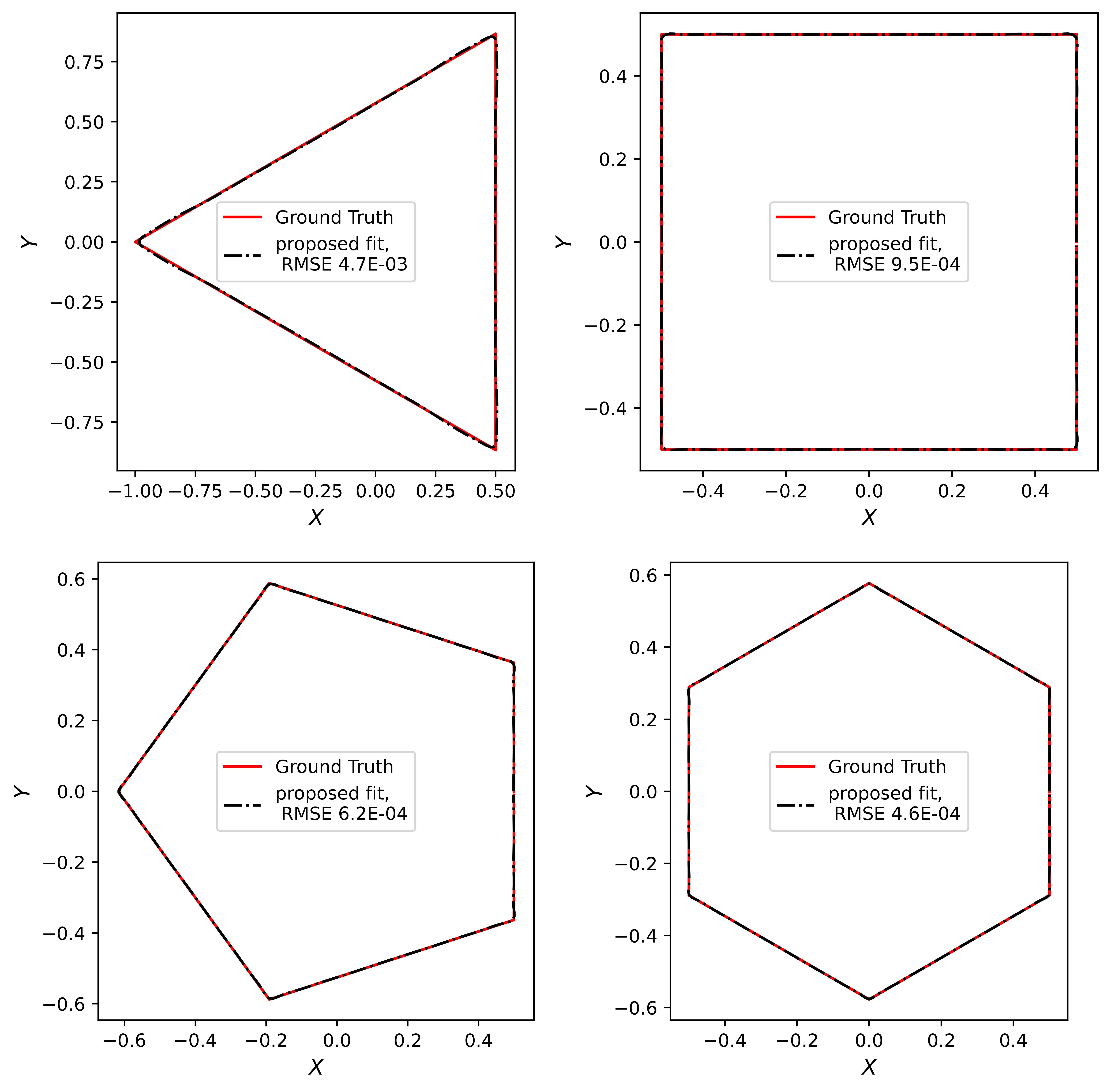}
	\caption[Fig. 1]{Fitness functions for unit triangle, square, pentagon, hexagon, and ellipse based on Eqs. (\ref{pxy}) with RMSE of fitting between the fitness functions and the unit shapes}
	\label{fig1}
\end{figure}

Inspired by the Fourier series, we formulated a high-precision trigonometric shape fitting function. We define the 2D regular polygon inscribed by a circle with a diameter of 1 as a unit polygon. Then, the fitness function for a unit polygon with E sides is given by:
\begin{equation}\label{pxy} 
\begin{split}
p_x(\phi)=S(\cos\phi(a+\frac{b}{\cos(E\phi)+1.08}-\frac{c}{\cos(2E\phi)+2})-d\cos((E-1)\phi))\\
p_y(\phi)=S(\sin\phi(a+\frac{b}{\cos(E\phi)+1.08}-\frac{c}{\cos(2E\phi)+2})+d\sin((E-1)\phi))
\end{split}
\end{equation}

where the values of parameters a, b, c, d, and S depend on the number of sides E and are listed in Table \ref{table1}. The function approximates the unit shape precisely with an angle parameter $\phi$. Thus, we can implement a simple and computationally efficient least squares algorithm to fit 2D shapes of any position and size, without any constraint conditions. Fig. \ref{fig1} show the fitness functions for a unit triangle, square, pentagon, and hexagon with calculated root-mean-square error (RMSE). 

\begin{table}[h!]
	\begin{center}
		\caption{Values of parameters a, b, c, d, and S for unit polygons and ellipse in Eqs. (\ref{pxy})}
		\label{table1}
		\begin{tabular}{cccccc}
			\hline
			Unit shape & a & b & c & d & S \\ \hline
			Triangle   & 17 & 0.14 & 0.42 & 5 & 1/24 \\
			Square     & 15 & 0.105 & 0.21 & 2 & 1/26 \\
			Pentagon   & 13 & 0.065 & 0.13 & 1 & 1/24 \\
			Hexagon    & 10 & 0.0375 & 0.0675 & 0.5 & 1/19 \\
			Circle     & 1  & 0  & 0  & 0  &  1/2 \\ \hline
		\end{tabular}
	\end{center}
\end{table}

\subsection{Shape parameters and error function}
\label{ShapeParaErrFunc}
Based on the unit shape presented in Eq. \ref{pxy}, we can parameterise corresponding symmetrical and asymmetrical shapes with a vector of six parameters represented by:
\begin{equation}\label{V}
	\mathbf{V} =\left[
	\begin{matrix}
		x_c & y_c & m_x & m_y & \alpha & \lambda\\
	\end{matrix}
	\right]^T
\end{equation}
where $x_c$ and $y_c$ are the coordinates of centroid, $m_x$ and $m_y$ are scaling factor to stretch the unit shape along the major and minor axes, $\alpha$ is a counterclockwise rotation angle, and $\lambda$ is the shear factor. 

Then we can fit polygons or ellipses to the data points by minimizing the error function given by:
\begin{equation}\label{L}
	\begin{split}
		&\mathbf{L}=\left[\begin{matrix} 
			\mathbf{L}_1 & \mathbf{L}_2 & \cdots & \mathbf{L}_n 
		\end{matrix}\right]^T,\\
		\mathbf{L}_i & = \mathbf{R}^{T}(\mathbf{X}_i-\mathbf{T})-\mathbf{M}\cdot \mathbf{P}_i
	\end{split}
\end{equation}

where $\mathbf{R}$ is the rotation matrix corresponding to $\alpha$, $\mathbf{X}_i$=[$x_i$, $y_i$]$^T$ are the coordinates of data points, $\mathbf{T}$=[$x_c$, $y_c$]$^T$ is the translation vector, and $\mathbf{P}_i$=[$p_x$, $p_y$]$^T$ is the unit shape defined in Eqs. (\ref{pxy}). $\mathbf{M}$ is defined as follows:
\begin{equation}\label{M}
	\begin{split}
		\mathbf{M}=
		\left[\begin{matrix} 
			m_x & 0 \\ \lambda m_y & m_y
		\end{matrix}\right]
	\end{split}
\end{equation}

\subsection{Estimation of the angle parameter $\phi$}
\label{EstPhi}
To fit the shape using least squares adjustment, a key step is to calculate the intermediate parameter $\phi$ in Eq. \ref{pxy} for each data point. A solution of $\phi$ can be computed for polygon and ellipse through Eqs. \ref{r} to \ref{phi_ellipse} . \par

We denote $r={p_y}/{p_x}$, and set $\theta=\arctan((y_i-y_c)/(x_i-x_c))-\alpha$. From Eq. \ref{L} we can have:
\begin{equation}\label{r}
	r=\frac{p_y}{p_x}=-\lambda+\tan\theta\frac{m_x}{m_y}\\
\end{equation}

We denote $w={a}/{d}$, where $a$ and $d$ are the parameters of the fitness function in Eq. \ref{pxy}, and use shorthand notations $\mathbf{c}=\cos(\phi)$ and $\mathbf{c}_n=\cos(n\phi)$. From Eq. \ref{pxy} we can get an approximation of $\cos(\phi)$ for polygon as follows:
\begin{equation}\label{cosphi}
	\begin{split}
		(r^2+1)(w^2\mathbf{c}^2+\mathbf{c}^2_{E-1}-2w\mathbf{c}\mathbf{c}_{E-1})+2w\mathbf{c}_E-w^2-1=0
	\end{split}
\end{equation}
where $E$ is the number of sides of the corresponding unit polygon. From Eq. \ref{cosphi}, we can always find two real solutions, $c_0$ and $c_1$, for $\cos\phi$. We assume $c_0 < c_1$ and the correct solution $c_t$ can be selected based on $\theta$ by:
\begin{equation}\label{ct}
	\begin{split}
		c_t=
		\begin{cases}
			c_0,& \text{for $\cos\theta<0$}\\
			c_1,& \text{otherwise}
		\end{cases}
	\end{split}
\end{equation}
Taking shear factor $\lambda$ in Eq. \ref{L} into consideration, $\phi$ can be calculated by taking the inverse cosine of $c_t$ and adjusting the sign with Eq. \ref{phi_polygon} for polygon.
\begin{equation}\label{phi_polygon}
	\phi= \text{sgn}(\sin(\theta - \arctan(\frac{\lambda m_y}{m_x}))) \arccos(c_t)
\end{equation}
where sgn($\cdot$) is the sign function. \par

For ellipse, $\phi$ can be calculated with Eq. \ref{phi_ellipse}.
\begin{equation}\label{phi_ellipse}
	\phi=\arctan(\frac{m_x}{m_y}\tan\theta)+\text{sgn}(\sin\theta)\text{stp}(\cos\theta)\pi
\end{equation}
where stp($\cdot$) is a step function defined as:
\begin{align}\label{stp}
	\text{stp}(x)= \begin{cases}
		0; \text{if } x \geq 0\\
		1; \text{if } x < 0
	\end{cases}
\end{align}

In practical implementation, we can simplify the computation of the angle parameter $\phi$ based on the above equations. Taking rectangle fitting as an example, setting $w=a/d=7.5$ from Table \ref{table1}, Eq. \ref{cosphi} gives:
\begin{subequations}\label{phi_rect:main}
\begin{equation}\label{cosphi_rect}
	16(r^2+1)\mathbf{c}^6-(84r^2-36)\mathbf{c}^4+(110.25r^2-9.75)\mathbf{c}^2-42.25=0
\end{equation}
Taking $\lambda=0$ into Eq. \ref{r} and setting $k=1/(r^2+1)$, Eq. \ref{cosphi_rect} can be reduced to:
\begin{equation}\label{cosphi_rect2}
	\mathbf{c}^6+(7.5k-5.25)\mathbf{c}^4+(6.890625-7.5k)\mathbf{c}^2- 2.640625k=0
\end{equation}
From Eq. \ref{cosphi_rect2} we can get:
\begin{equation}\label{cosphi_rect3}
	c_1=-c_0=\sqrt{1.75-2.5k-(q_1+q_2)^{1/3}-(q_1-q_2)^{1/3}}
\end{equation}
where
\begin{align}\label{phi_rectangle:q1q2}
	\begin{split}
		q_1=p_1(k-0.5)^3-p_2(k-0.5),\\
		q_2=p_3 \sqrt{(k^2-k)(k -k^2-p_4)}
	\end{split}
\end{align}
where $p_1$=15.625, $p_2$=5.24609375, $p_3$=8.3998, $p_4$=0.2786875. 
As $c_1=-c_0$, from Eq. \ref{phi_polygon}, the value of $\phi$ for rectangle can be calculated as follows:
\begin{align}\label{phi_rectangle}
	\begin{split}
		\phi=&\text{sgn}(\tan\theta)\arccos(c_1)+\text{sgn}(\sin\theta)\text{stp}(\cos\theta)\pi
	\end{split}
\end{align}
\end{subequations}

\subsection{Least squares adjustment}
\label{lsadjust}
With determined shape parameters and error functions described in Section \ref{ShapeParaErrFunc} and computed $\phi$ described in Section \ref{EstPhi}, the standard Gauss-Newton method is used to minimise $\mathbf{L}$:
\begin{subequations}
	\begin{align}\label{g}
		\mathbf{g}=-(\mathbf{A}^T\mathbf{A})^{-1}\mathbf{A}^T\mathbf{L}
	\end{align}
	\begin{align}
		\mathbf{V}:=\mathbf{V}+\mathbf{g}
	\end{align}
\end{subequations}
where $\mathbf{A}$ is partial derivatives of $\mathbf{L}$ given by:
\begin{subequations}\label{A:main}
	\begin{align}\tag{\ref{A:main}}
	\begin{split}
		&\mathbf{A}=\left[\begin{matrix} 
			\mathbf{A}_1 & \mathbf{A}_2 & \cdots & \mathbf{A}_n 
		\end{matrix}\right]^T,\\
		&\mathbf{A}_i =\left[
		\begin{matrix}
			\frac{\partial{\mathbf{L}_i}}{\partial{x_c}} & 	\frac{\partial{\mathbf{L}_i}}{\partial{y_c}} & \frac{\partial{\mathbf{L}_i}}{\partial{m_x}} & \frac{\partial{\mathbf{L}_i}}{\partial{m_x}} & \frac{\partial{\mathbf{L}_i}}{\partial{\alpha}}&
			\frac{\partial{\mathbf{L}_i}}{\partial{\lambda}}
			\\\end{matrix}
		\right]
	\end{split}
	\end{align}
From Eq. (\ref{L}), we can have:
	\begin{align}\label{A:Li_xyc}
	\begin{split}
		&\frac{\partial{\mathbf{L}_i}}{\partial{x_c}}=
		\left[\begin{matrix}-\cos\alpha \\ \sin\alpha
		\end{matrix}\right] - \left[\begin{matrix} 
			m_x\frac{\partial{p_x}}{\partial x_c}\\ 
			m_y(\lambda\frac{\partial{p_x}}{\partial x_c}+\frac{\partial{p_y}}{\partial x_c})
		\end{matrix}\right],\\
		&\frac{\partial{\mathbf{L}_i}}{\partial{y_c}}=
		\left[\begin{matrix} -\sin\alpha \\ -\cos\alpha
		\end{matrix}\right] - \left[\begin{matrix} 
			m_x\frac{\partial{p_x}}{\partial y_c}\\ 
			m_y(\lambda\frac{\partial{p_x}}{\partial y_c}+\frac{\partial{p_y}}{\partial y_c})
		\end{matrix}\right]
	\end{split}
	\end{align}

	\begin{align}\label{A:Li_mxy}
		\begin{split}
		\frac{\partial{\mathbf{L}_i}}{\partial{m_x}}=- \left[\begin{matrix} 
			p_x+m_x\frac{\partial{p_x}}{\partial m_x}\\ m_y(\lambda\frac{\partial{p_x}}{\partial m_x}+\frac{\partial{p_y}}{\partial m_x})
		\end{matrix}\right],\\
		\frac{\partial{\mathbf{L}_i}}{\partial{m_y}}=- 
		\left[\begin{matrix}
			m_x\frac{\partial{p_x}}{\partial m_y}\\
			\lambda p_x+p_y+m_y(\lambda\frac{\partial{p_x}}{\partial m_y}+\frac{\partial{p_y}}{\partial m_y})
			\end{matrix}\right]
		\end{split}
	\end{align}

	\begin{align}\label{A:Li_alpha}
	\frac{\partial{\mathbf{L}_i}}{\partial{\alpha}}= \left[\begin{matrix} 
		-\sin\alpha(x_i-x_c)+\cos\alpha(y_i-y_c) \\ -\cos\alpha(x_i-x_c)-\sin\alpha(y_i-y_c)
	\end{matrix}\right] - 
	\left[\begin{matrix} 
		m_x\frac{\partial{p_x}}{\partial \alpha}\\
		m_y\frac{\partial{p_y}}{\partial \alpha}
	\end{matrix}\right]
	\end{align}
	
	\begin{align}\label{A:Li_lambda}
		\frac{\partial{\mathbf{L}_i}}{\partial{\lambda}}=- \left[\begin{matrix} 
			m_x\frac{\partial{p_x}}{\partial \lambda}
			\\ m_y(p_x+\lambda\frac{\partial{p_x}}{\partial \lambda}+\frac{\partial{p_y}}{\partial \lambda})
		\end{matrix}\right]
	\end{align}
\end{subequations}

\subsection{Algorithm Implementation}\label{implement}
Before applying the proposed method, the data should be preprocessed with the following steps, which are analogous to the data preprocessing in circle fitting \cite{Abdul-Rahman}:\par
a) Translate the data points to the origin by subtracting their mean values: $x'_i=x_i-\overline{x}$ and $y'_i=y_i-\overline{y}$, where $\overline{x}$ and $\overline{y}$ are sample means of  $x_i$ and $y_i$, respectively.\par
b) Normalize the data by dividing $x'_i$ and $y'_i$ by the root mean squared distance, $d_{\text{rms}}$, of the translated data points from the origin.\par
c) For shapes that are elongated or flattened, we recommend to apply singular value decomposition (SVD) to the translated and normalized data points $\mathbf{usv}^T = \mathbf{X}'$, and obtain the angle of the principal axis as the initial estimate of $\alpha$. \par

These steps can reduce rounding error and mis-convergence in iteration. We can start the search of the parameters  ($x_c$, $y_c$, $m_x$, $m_y$, $\lambda$) at (0, 0, 1, 1, 0) with the calculated initial value of $\alpha$. To simplify the fitting problem and reduce the computational cost, we may adjust the vector based on the degrees of freedom of the fitted shape. For example, if the fitted shape is a rectangle or an ellipse, the number of parameters in Eq. (\ref{V}) is reduced to five as $\lambda$ is fixed to 0. If the fitted shape is a right triangle, then we can fix $\lambda=\sqrt{3}/3$. The optimal parameters after iteration are then transformed back by applying the inverse operations of scaling and translation with Eqs. \ref{scale_back}.

\begin{equation}\label{scale_back}
	\begin{aligned}
		x_c \gets x_c  d_{\text{rms}} + \overline{x}, 
		y_c \gets y_c  d_{\text{rms}} + \overline{y}, \\
		m_x \gets m_x  d_{\text{rms}}, 
		m_y \gets m_y  d_{\text{rms}}
	\end{aligned}        
\end{equation}

Algorithm \ref{algorithm} shows the procedures of data preprocessing and iterative least squares adjustment. 

\begin{algorithm}
	\caption{Proposed Shape Fit Algorithm}\label{algorithm}
	\begin{algorithmic}[1]
		\Procedure{Data preprocessing}{}
			\For{each ($x_i$, $y_i$) in $\mathbf{X}$}\par
			\hskip\algorithmicindent $x_i \gets (x_i-\overline{x})/d_{\text{rms}}$\par
			\hskip\algorithmicindent $y_i \gets (y_i-\overline{y})/d_{\text{rms}}$
			\EndFor
			\State Compute $\alpha_0$ by $\mathbf{u}, \mathbf{s}, \mathbf{v}^T := \text{svd}({\mathbf{X}})$ \par
			\hskip\algorithmicindent $\alpha_0 := \arctan(\mathbf{v}^T)$
		\EndProcedure
		\Procedure{Iterative least squares minimization}{}
			\State Initialise $\mathbf{V} := [0, 0, 1, 1, \alpha_0, 0]^T$, $\epsilon := 10^{-6}$
			\While {not converged} 
			\State Compute $\phi$ by Eq. (\ref{phi_polygon}) for polygon or Eq. (\ref{phi_ellipse}) for ellipse
			\State Compute $\mathbf{L}$ by Eq. (\ref{L})
			\State Compute $\mathbf{A}$ by Eq. (\ref{A:main})
			\State Compute $\mathbf{g}$ by Eq. (\ref{g})
			\If {${\|\mathbf{g}\|}<\epsilon \|\mathbf{V}\|$} break 
			\Else
			\State $\mathbf{V} \gets \mathbf{V}+\mathbf{g}$
			\EndIf
			\EndWhile
		\EndProcedure
		\State restore ($x_c, y_c, m_x, m_y$) in estimated $\mathbf{V}$ by Eqs. (\ref{scale_back})
	\end{algorithmic}
\end{algorithm}

\section{Validation}\label{validation}
\subsection{Data Description}
\label{DataDescription}
We designed four tests to validate our proposed method with different types of data and shapes. Tests 1 and 2 used synthetic data of eleven shapes with varying parameters shown in Table \ref{table2}. The data points used in Test 1 were corrupted by Gaussian noise with a standard deviation of 0.5 to simulate measurement errors. The data points used in Test 2 were noise-free but partially missing. We deliberately removed some points near vertices of polygons and some points on one side of each ellipse to create gaps in the data. This way, we can evaluate the accuracy and robustness of our proposed method to noisy and incomplete data. We also demonstrate that our method does not rely on the vertices of polygons.

\begin{figure}[h!]
	\centering
	\includegraphics[width=1\linewidth]{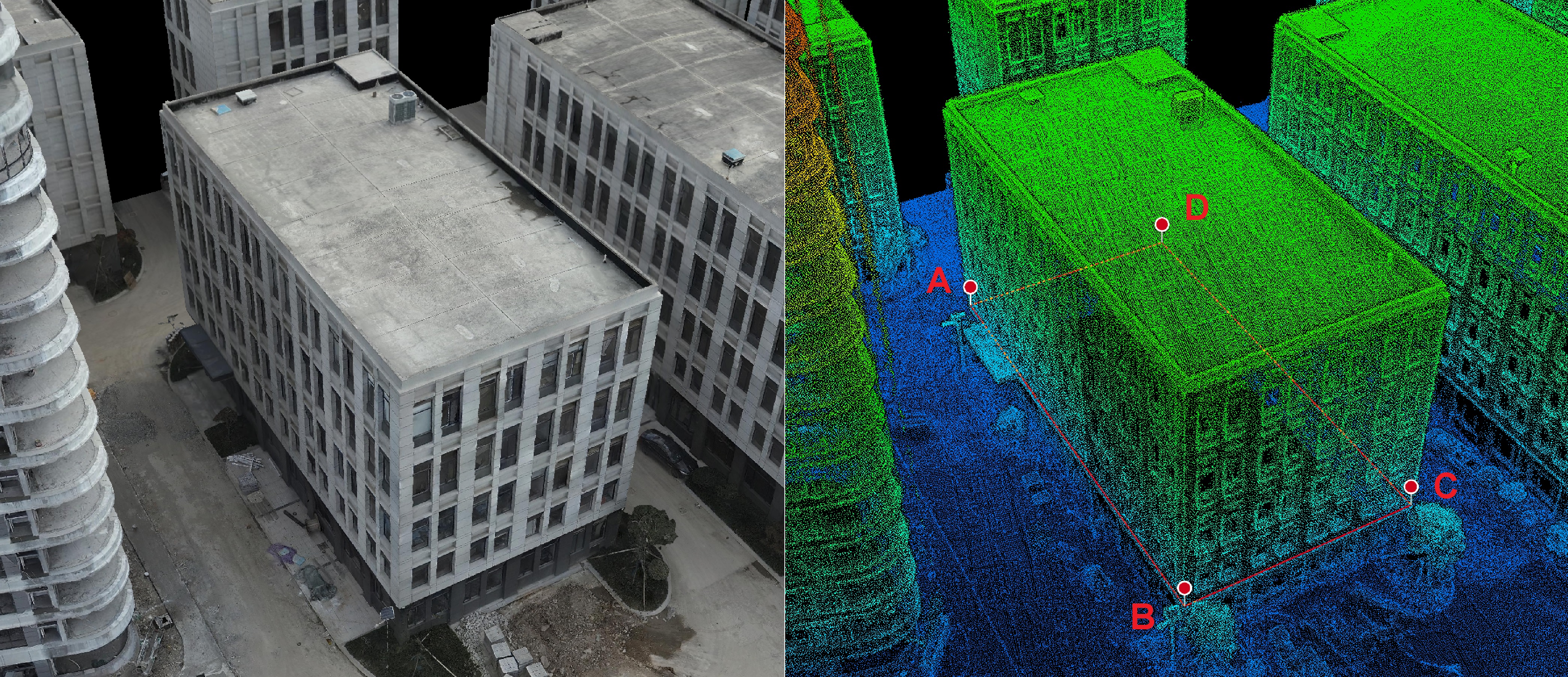}
	\caption[Fig. 2]{Point cloud of a building from aerial photogrammetry and its 3D model. (a) 3D model of the building. (b) Point cloud of the building. The data used in Test 3 is the sliced point cloud near $P_{\text{ABCD}}$, which is a plane defined by four corner points of the building.}
	
	\label{fig2}
\end{figure}

Tests 3 and 4 used real data with rectangular and elliptical shapes because they are ubiquitous in the real world and have many applications in computer vision, image processing, and remote sensing. Test 3 used a point cloud of a building obtained by aerial photogrammetry (Fig. \ref{fig2}), which can be approximated by a rectangle. The data was collected by flying a DJI M300 RTK drone equipped with a Zenmuse P1 photogrammetry camera over the building. The true parameters of the rectangle in Test 3 were computed using the planar coordinates of the four corner points of the building. The true coordinates of the building corners were derived from the as-built drawing, which was created based on a total station survey of the building. Test 4 used an image of a shape matching toy with polygonal and circular shapes to demonstrate the versatility of our proposed method. The image was preprocessed by converting it to gray scale, applying a Gaussian filter for smoothing, and extracting shape contours using Canny edge detector before applying our proposed shape fitting method.

In the tests, we compared our proposed method with two existing methods for rectangle and ellipse fitting proposed by Sinnreich \cite{Sinnreich} and Prasad \cite{Prasad}, respectively, when a ground truth reference is available. We used root-mean-square error (RMSE) as the metric to evaluate fitting accuracy of the proposed and referenced methods. 

\begin{table}[h!]
	\begin{center}
		\caption{Parameters of eleven shapes in Tests 1 and 2}
		\label{table2}
		\begin{tabular}{cccccccc}
			\hline
			Shape No. & $x_c$	& $y_c$ & $m_x$ & $m_y$ & $\alpha$/deg & $\lambda$   & Type	\\\hline
			T0   	& 4 	& 43 	& 15 	& 15 	& 35	& /   		  	& regular triangle \\
			T1   	& 0 	& 17 	& 28 	& 9 	& 100	& /   		  	& isoceles triangle\\
			T2   	& 15 	& -4 	& 15 	& 9 	& 270	& $\sqrt{3}/3$	& right triangle \\
			R0   	& 32 	& 39 	& 19 	& 19 	& -10	& /   		  	& square 	  \\
			R1   	& 33 	& 4 	& 8 	& 41 	& 50	& /				& rectangle   \\ 
			R2   	& 32 	& 21 	& 8 	& 16 	& 80	& $\sqrt{3}/2$	& rhombus \\ 
			R3   	& 49 	& 27 	& 15 	& 7 	& 60	& 0.8			& parallelogram \\ 
			H0   	& 61 	& 47 	& 20 	& 20 	& 15	& /				& regular hexagon  \\ 
			H1   	& 68 	& 24 	& 24 	& 16 	& 10	& 0.6			& hexagon \\ 
			E0   	& 52 	& 8 	& 8 	& 12 	& 20	& /				& ellipse \\ 
			E1   	& 68 	& 0 	& 32 	& 8 	& 55	& /				& ellipse \\ \hline
		\end{tabular}
	\end{center}
\end{table}

\subsection{Results}\label{result}
\begin{figure}[h!]
	\centering
	\includegraphics[width=1\linewidth]{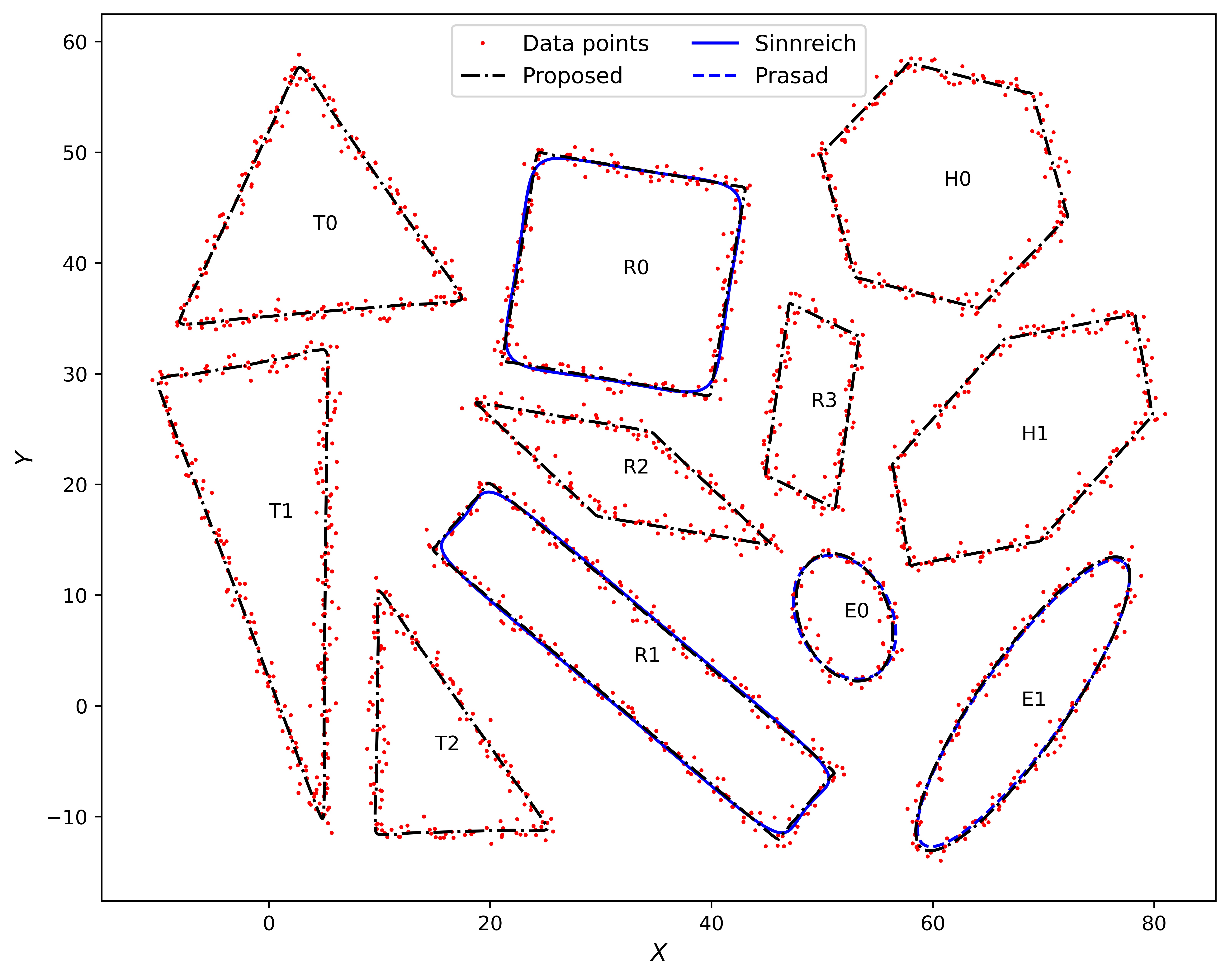}
	\caption[Fig. 3]{Shape fitting results for noisy points Using proposed and referenced methods in Test 1}
	\label{fig3}
\end{figure}

Test 1 evaluated the accuracy and robustness of our proposed method for fitting geometric shapes to noisy data. Fig. \ref{fig3} shows that our proposed method and the referenced methods can fit the noisy data points with low errors, ranging from 0.5 to 1.3 in RMSE. The estimated parameters and RMSE of fitting points in Test 1 are shown in Table \ref{table3}. The results show that the RMSE is higher for elongated or flattened shapes such as T1 and R1 because the estimation of a short edge of these shapes is more sensitive to noise. The results also show that our proposed method has similar RMSE as Sinnreich's method \cite{Sinnreich} for rectangle fitting (R0 and R1) and slightly lower RMSE than Prasad's method \cite{Prasad} for ellipse fitting (E0 and E1). The test results indicate that our method is competitive with the referenced methods, considering that the proposed method is more general and versatile.

\begin{table}[h!]
	\begin{center}
	\caption{Shape parameter estimation and RMSE for various shapes and methods in Test 1.}
	\label{table3}
	\begin{tabular}{ccccccccc}
		\hline
		Shape   & Method	& $x_c$	& $y_c$ & $m_x$ & $m_y$ & $\alpha$/deg & $\lambda$  & RMSE	\\ \hline
		T0 & Proposed  & 4.069 & 43.017 & 14.964 & 14.959 & 154.88 & /  & 0.639\\
		T1 & Proposed  & 0.023 & 16.989 & 28.042 & 9.079 & 100.08 & /  & 1.122\\
		T2 & Proposed  & 14.963 & -4.053 & 8.701 & 15.387 & 35.32 & 0.199 & 0.738\\
		R0 & Sinnreich  & 32.038 & 38.918 & 19.301 & 19.158 & -8.90 & /  & 0.604\\
		R0 & Proposed  & 32.070 & 38.995 & 19.109 & 19.065 & 80.39 & /  & 0.550\\
		R1 & Sinnreich  & 33.109 & 3.931 & 40.605 & 8.228 & 140.08 & /  & 1.295\\
		R1 & Proposed  & 32.958 & 4.046 & 40.731 & 7.998 & 140.01 & /  & 1.316\\
		R2 & Proposed  & 32.081 & 20.976 & 8.940 & 15.070 & 46.89 & -0.889 & 1.055\\
		R3 & Proposed  & 49.098 & 27.118 & 6.673 & 15.685 & -7.92 & -0.136 & 0.706\\
		H0 & Proposed  & 61.015 & 47.000 & 19.945 & 19.973 & 16.01 & /  & 0.503\\
		H1 & Proposed  & 68.095 & 24.026 & 23.918 & 16.060 & 9.98 & 0.594 & 0.584\\
		E0 & Prasad  & 52.029 & 8.014 & 11.785 & 8.476 & -63.12 & /  & 0.506\\
		E0 & Proposed  & 52.011 & 8.008 & 11.985 & 8.094 & 111.83 & /  & 0.475\\
		E1 & Prasad  & 68.173 & 0.245 & 31.249 & 7.846 & 54.71 & /  & 0.781\\
		E1 & Proposed  & 68.128 & 0.194 & 31.841 & 8.043 & 55.20 & /  & 0.739\\ \hline
	\end{tabular}
	\end{center}
\end{table}

\begin{figure}[h!]
	\centering
	\includegraphics[width=1\linewidth]{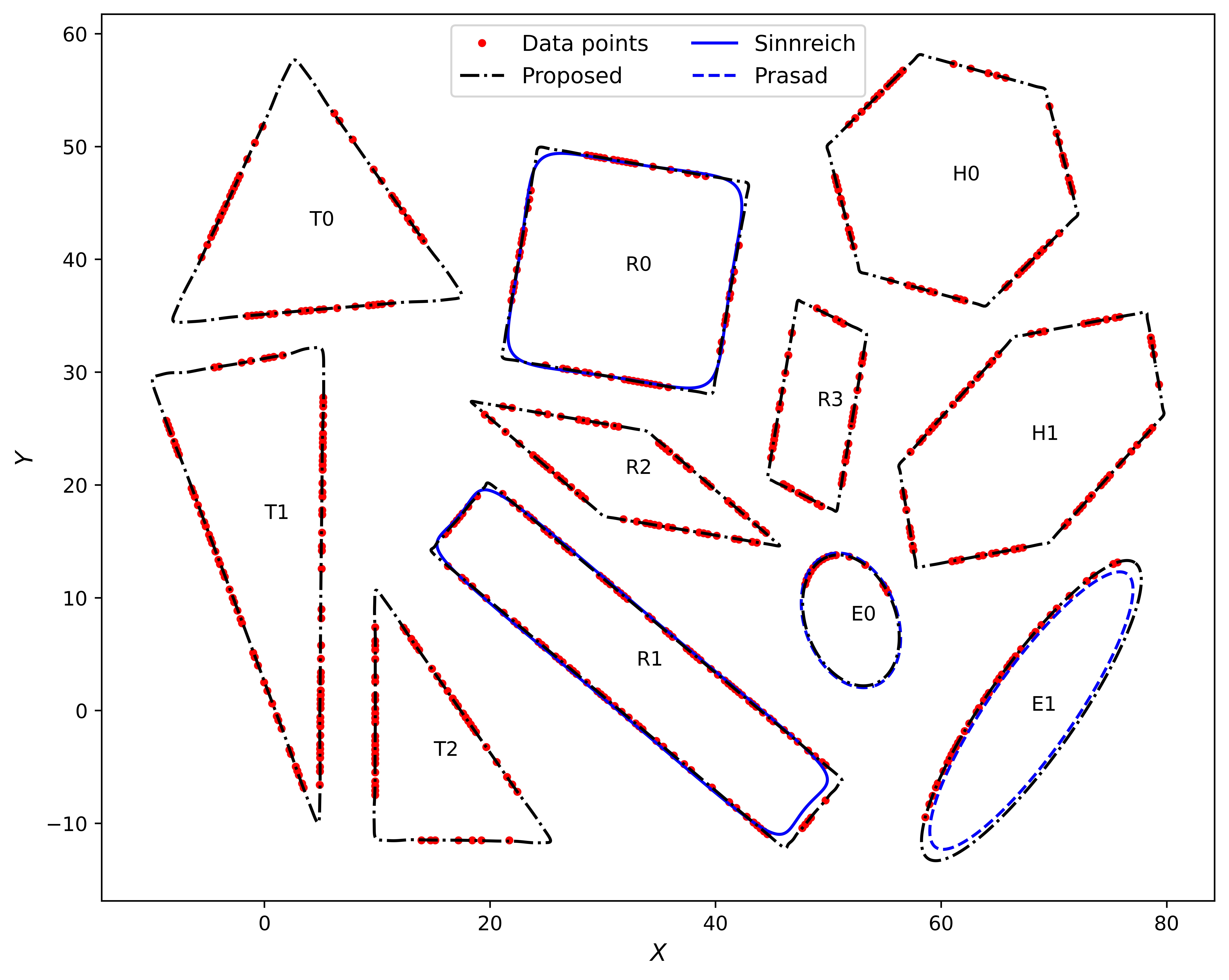}
	\caption[Fig. 4]{Shape fitting results for noise-free and incomplete points using proposed and referenced methods in Test 2.}
	\label{fig4}
\end{figure}

Test 2 evaluated the accuracy and robustness of our proposed method for fitting geometric shapes to incomplete data. Fig. \ref{fig4} illustrates the fitting shapes obtained by the proposed method and referenced methods in Test 2. As shown in Fig. \ref{fig4}, we can see that Sinnreich's fit of R1 and Prasad's fit of E1 are slightly smaller than the true shapes, which indicates an underfitting problem. In contrast, our proposed fit matches the data points more closely, which indicates a better fitting accuracy. Table \ref{table4} presents the estimated parameters and RMSE of fitting points in Test 2. The RMSE values of our proposed method range from 0.009 to 0.074 for polygons, which are lower by 80\% and 93\% than Sinnreich's fit. This indicates that the proposed method can handle incomplete data more effectively. For ellipses, the RMSE of the proposed method is exactly zero for these test cases, which indicates that the proposed method can recover the true parameters of the ellipse from incomplete data. Prasad's fit estimates the centroid and orientation of ellipse correctly, but has a slight error in estimating the major and minor axes of ellipse with incomplete data. This error results in an underfitting problem, as shown in Fig. \ref{fig4}. 

It should be noted that some estimated parameters in Table \ref{table4} differ from the given parameters in Table \ref{table2}, even though the corresponding geometric shapes fit well with the data points. For example, Shape R3 has given parameters ($x_c$, $y_c$, $m_x$, $m_y$, $\alpha$, $\lambda$) = (49, 27, 15, 7, 60, 0.8) in Table \ref{table2} and estimated parameters ($x_c$, $y_c$, $m_x$, $m_y$, $\alpha$, $\lambda$) = (49.005, 27.018, 6.552, 16.040, -9.6, -0.117) in Table \ref{table4}. This discrepancy is caused by the fact that the parameter model used in Eq. \ref{V} corresponding to an asymmetrical shape has multiple solutions that can produce the same shape.

\begin{table}[h!]
	\begin{center}
	\caption{Shape parameter estimation and RMSE for various shapes and methods in Test 2.}
	\label{table4}
	\begin{tabular}{ccccccccc}
		\hline
		Shape 	& Method	& $x_c$	& $y_c$ & $m_x$ & $m_y$ & $\alpha$/deg & $\lambda$   & RMSE	\\ \hline
		T0 & Proposed  & 4.010 & 42.982 & 15.082 & 14.979 & 155.09 & /  & 0.039\\
		T1 & Proposed  & -0.016 & 17.087 & 27.912 & 8.989 & 100.01 & /  & 0.074\\
		T2 & Proposed  & 15.043 & -4.053 & 8.538 & 15.883 & 34.84 & 0.191 & 0.043\\
		R0 & Sinnreich & 31.958 & 38.996 & 18.872 & 18.957 & -8.61 & /  & 0.102\\
		R0 & Proposed  & 31.993 & 39.009 & 19.017 & 18.988 & 80.25 & /  & 0.020\\
		R1 & Sinnreich & 32.602 & 4.303 & 40.171 & 8.196 & 139.95 & /  & 0.344\\
		R1 & Proposed  & 32.996 & 4.002 & 41.017 & 8.003 & 140.00 & /  & 0.023\\
		R2 & Proposed  & 32.002 & 21.003 & 8.241 & 15.656 & 49.23 & -0.870 & 0.074\\
		R3 & Proposed  & 49.005 & 27.018 & 6.552 & 16.040 & -9.60 & -0.117 & 0.066\\
		H0 & Proposed  & 61.001 & 47.003 & 20.002 & 20.000 & 14.68 & /  & 0.018\\
		H1 & Proposed  & 68.000 & 24.001 & 23.935 & 16.035 & 9.73 & 0.603 & 0.009\\
		E0 & Prasad  & 52.000 & 8.000 & 12.346 & 8.231 & -70.00 & /  & 0.157\\
		E0 & Proposed  & 52.000 & 8.000 & 12.000 & 8.000 & 110.00 & /  & 0.000\\
		E1 & Prasad  & 68.000 & -0.000 & 29.583 & 7.396 & 55.00 & /  & 0.643\\
		E1 & Proposed  & 68.000 & -0.000 & 32.000 & 8.000 & 55.00 & /  & 0.000\\ \hline
	\end{tabular}
	\end{center}
\end{table}

The test result using real point cloud data indicates that the proposed method can achieve centimetre level accuracy in high-precision applications. Table \ref{table5} shows the estimated parameters and RMSE of fitting points in Test 3. From Table \ref{table5}, we can observe that the estimated parameters of the proposed fit are close to the ground truth, whilst Sinnreich's fit overestimates the length of the short edge of the rectangle. Fig. \ref{fig3} illustrates the fitted shapes by both methods overlaid on the point cloud data. A close look at Fig. \ref{fig3} shows that Sinnreich's fit exhibits a slight outward deviation from the long edge of the rectangle except near the vertices, where it curves inward. However, the proposed fit shows closer proximity to the rectangle, especially in capturing the vertices.

\begin{table}[h!]
	\begin{center}
		\caption{Shape parameter estimation and RMSE for the rectangular building in Test 3}
		\label{table5}
		\begin{tabular}{cccc}
			\hline
			Parameters  		& Ground truth & Sinnreich fit  & proposed fit   \\ \hline
			$x_c$/m             & 295595.930   & 295595.883     & 295595.947     \\
			$y_c$/m             & 95593.943    & 95594.002      & 95593.912      \\
			$m_x$/m             & 33.100       & 32.950         & 33.112         \\
			$m_y$/m             & 16.685       & 17.645         & 16.743         \\
			$\alpha$/deg        & 155.77       & 156.24         & 155.86        \\ \hline
			\textbf{RMSE/m} 	&              & \textbf{0.514} & \textbf{0.081} \\ \hline
		\end{tabular}
	\end{center}
\end{table}

\begin{figure}[h!]
	\centering
	\includegraphics[width=0.7\linewidth]{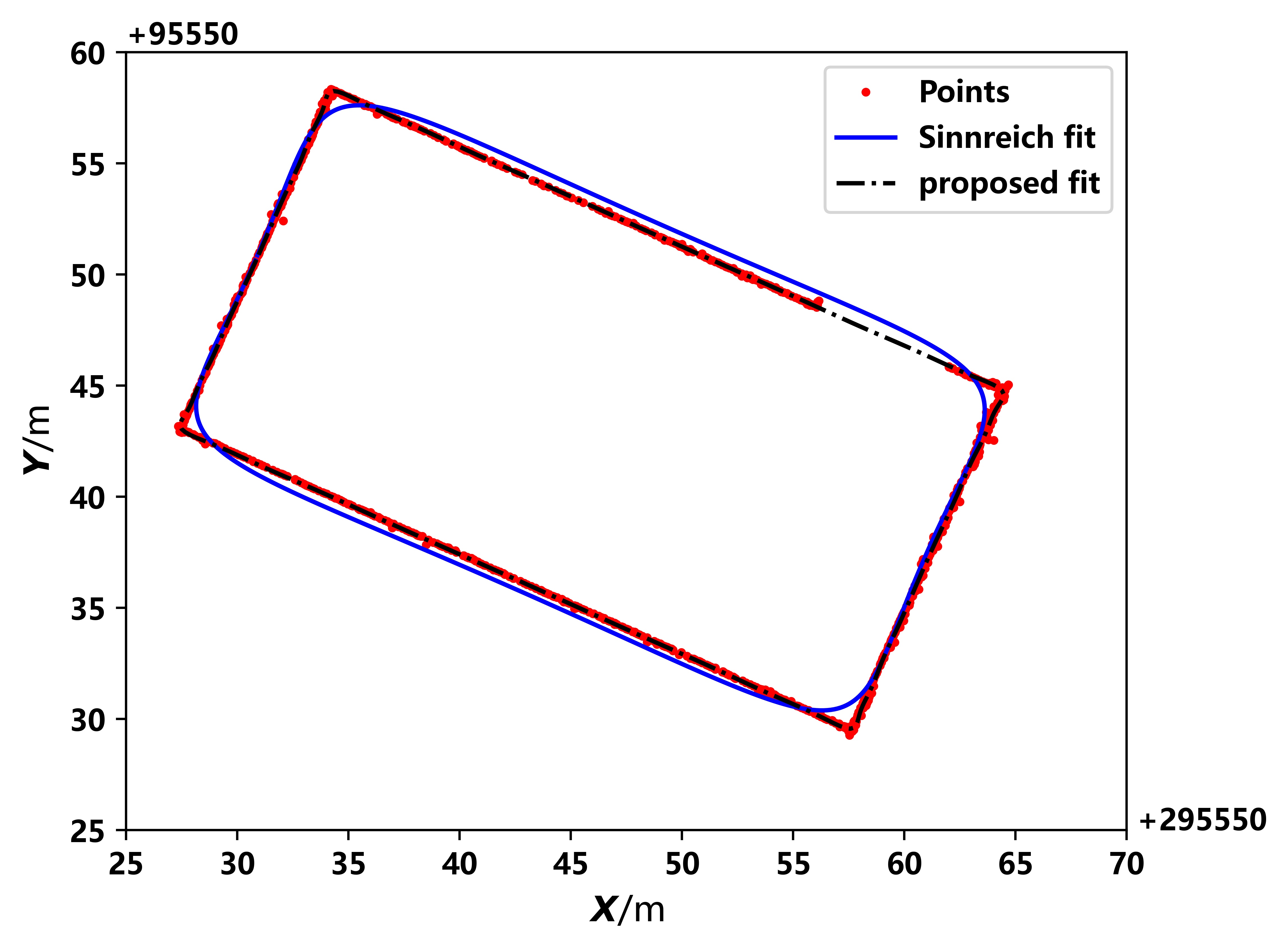}
	\caption[Fig. 5]{Shape fitting results for real point cloud data using proposed and referenced methods in Test 3.}
	\label{fig5}
\end{figure}

\begin{figure}[h!]
	\centering
	\begin{subfigure}{0.6\textwidth}
		\includegraphics[width=\textwidth]{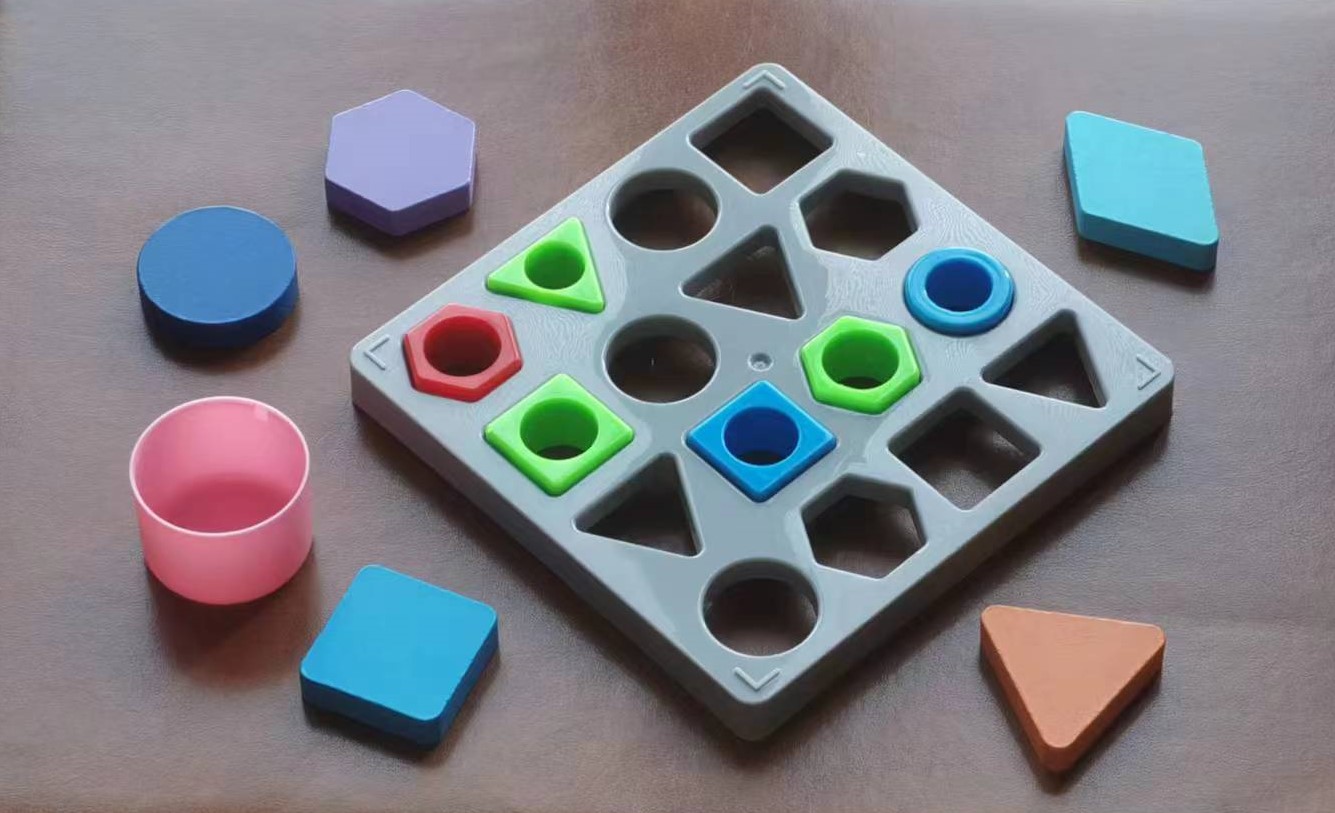}
		\caption{Original image}
		\label{fig6:a}
	\end{subfigure}
	\hfill
	\begin{subfigure}{0.45\textwidth}
		\includegraphics[width=\textwidth]{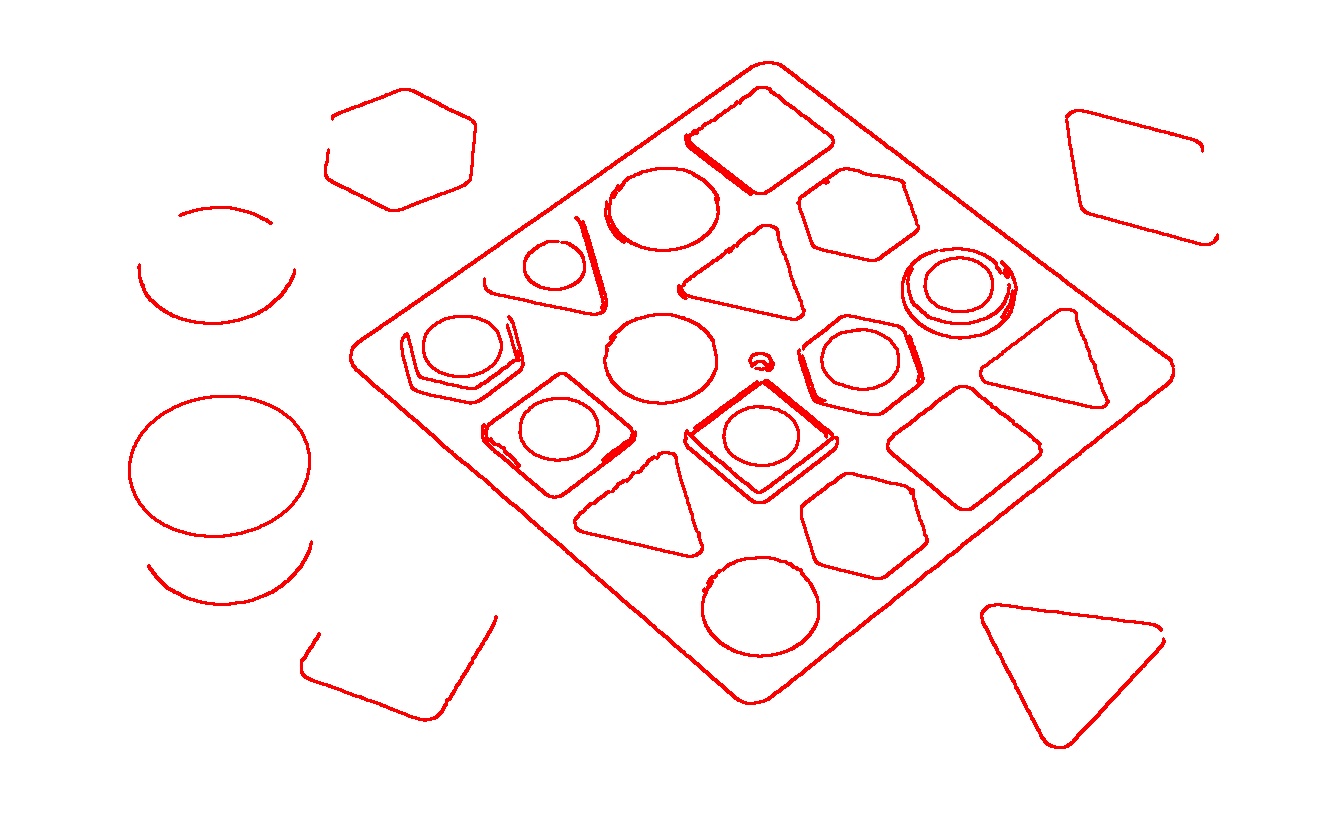}
		\caption{Input contours as input}
		\label{fig6:b}
	\end{subfigure}
	\hfill
	\begin{subfigure}{0.45\textwidth}
		\includegraphics[width=\textwidth]{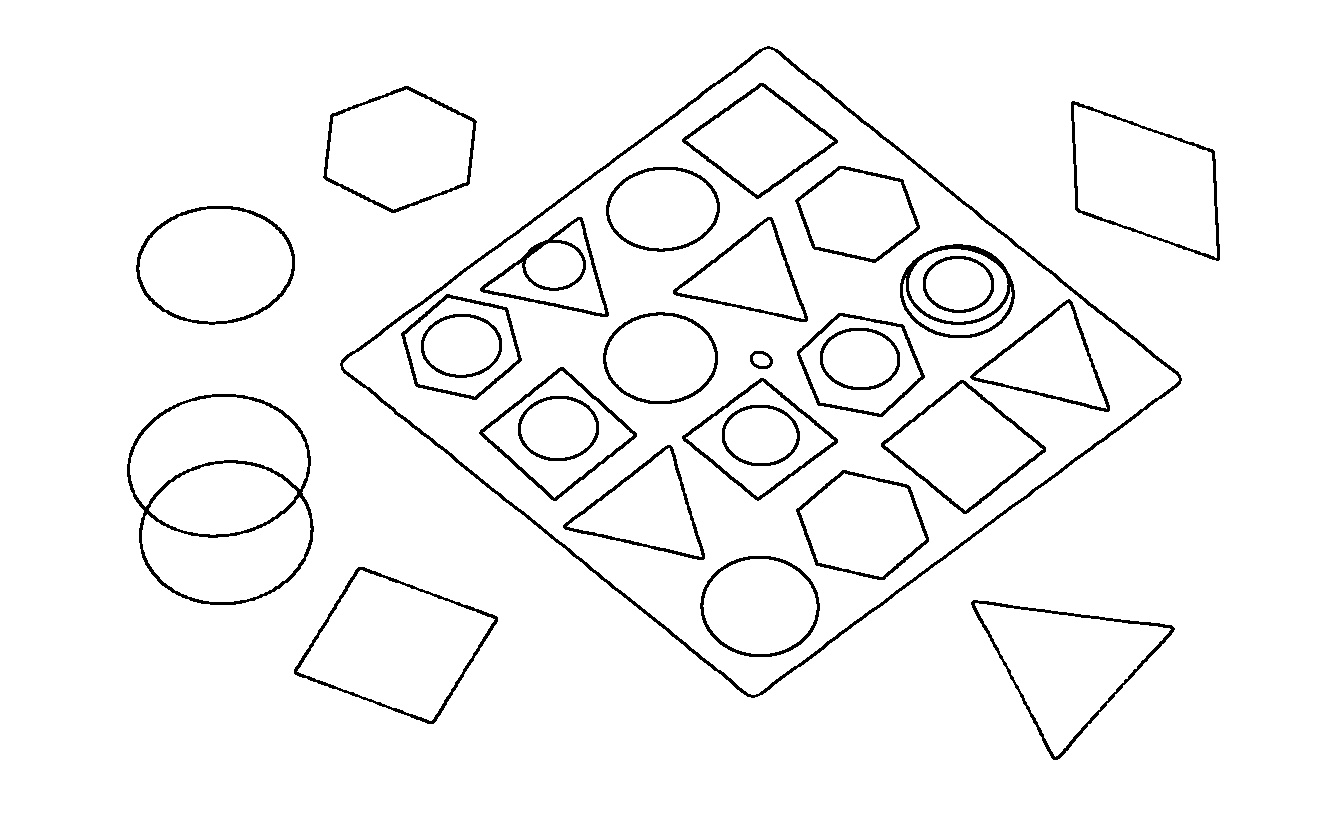}
		\caption{Fitting result of proposed method}
		\label{fig6:c}
	\end{subfigure}
	\caption{Example of a real image and the shapes fitted by the proposed method.}
	\label{fig:figures}
\end{figure}

Test 4 shows that our proposed method can fit various shapes in a real image of a shape-matching toy, which contains incomplete and noisy contours. From Fig. \ref{fig6:b}, we can see that some of the contours that belong to different shapes, such as the bottom-left square and top-right rhombus, are not complete. Some edges of shapes are not captured after the image pre-processing because of the changing light reflection along the shape edges in the original image (Fig. \ref{fig6:a}). However, the proposed method still works well in fitting all the shapes as shown in Fig. \ref{fig6:c}, demonstrating that our method is robust to contour gaps and noise, and versatile to polygonal and circular shapes.

\section{Conclusions}
\label{conclusions}
This work has presented a novel method for accurate shape fitting that can handle polygons and ellipses. We validated our method with simulated and real data. The simulated tests show that the proposed method is robust to noisy data and incomplete data. The real data tests further demonstrate the accuracy and versatility of the proposed method to the real-world point cloud data and image data. The proposed method was compared with the existing methods in the literature. The results show that the proposed method achieved higher accuracy and better versatility. The proposed method has potential applications in various fields that involve shape analysis, such as computer vision, image processing, and remote sensing.







\clearpage

\begin{thebibliography}{00}


\bibitem{Mai}
	F. Mai, Y.S. Hung, H. Zhong, W. F. Sze, 
	A hierarchical approach for fast and robust ellipse extraction, 
	Pattern Recognit. 41 (8) (2008) 2512-2524,
	doi:10.1016/j.patcog.2008.01.027.
\bibitem{Abdul-Rahman}
	H. Abdul-Rahman, N. Chernov,
	Fast and numerically stable circle fit, 
	J Math Imaging Vis 116 (2014) 289–295, 
	doi:10.1007/s10851-013-0461-4.
\bibitem{Prasad}	
	D.K. Prasad, M.K. Leung, C. Quek,
	ElliFit: An unconstrained, non iterative, least squares based geometric ellipse fitting method, 
	Pattern Recognit. 46 (5) (2013) 1449-1465,
	doi: 10.1016/j.patcog.2012.11.007.
\bibitem{Wang}
	T. Wang, Z. Shi, B. Yu,
	A parameterized geometric fitting method for ellipse, 
	Pattern Recognit. 116 (2021) 107934, 
	doi:10.1016/j.patcog.2021.107934.
\bibitem{Maalek}
	R. Maalek, D.D. Lichti,
	New confocal hyperbola-based ellipse fitting with applications to estimating parameters of mechanical pipes from point clouds, 
	Pattern Recognit. 116 (2021) 107948, 
	doi:10.1016/j.patcog.2021.107948.
\bibitem{Dong}
	H. Dong, E. Asadi, G. Sun, D.K. Prasad, I.-M. Chen,
	Real-time robotic manipulation of cylindrical objects in dynamic scenarios through elliptic shape primitives, IEEE Trans. Robot. (2018) 1–19,
	doi:10.1109/tro.2018.2868804.
\bibitem{Quan}
	Y. Quan, L. Lau,
	Development of a trajectory constrained rotating arm rig for testing GNSS kinematic positioning, 
	Measurement 140 (2019) 479-485, 
	doi:10.1016/j.measurement.2019.04.013.
\bibitem{Lu}
	Z. Lu, B. Liu, K. Zhang, H. Lin and Y. Zhang, 
	A Method for Measuring the Inclination of Forgings Based on an Improved Optimization Algorithm for Fitting Ellipses,
	IEEE Trans. Instrum. Meas. 72 (2023) 1-11,
	doi: 10.1109/TIM.2022.3221761.
\bibitem{LiJ}
	J. Li, Y. Wang, B. Lei, J.-Z. Cheng, J. Qin, T. Wang, S. Li, D. Ni, 
	Automatic fetal head circumference measurement in ultrasound using random forest and fast ellipse fitting, 
	IEEE J. Biomed. Health, 22 (1) (2018) 215-223, 
	doi: 10.1109/JBHI.2017.2703890.
\bibitem{Ranjbarzadeh}
	R. Ranjbarzadeh, S.B. Saadi, 
	Automated liver and tumor segmentation based on concave and convex points using fuzzy c-means and mean shift clustering, 
	Measurement 150 (2020) 107086,
	doi: 10.1016/j.measurement.2019.107086.
\bibitem{Yang}
	J. Yang, Z. Jiang,
	Rectangle fitting via quadratic programming, 
	in: 2015 IEEE 17th International Workshop on Multimedia Signal Processing, Xiamen, China, 2015, pp. 1-6, 
	doi: 10.1109/MMSP.2015.7340875.
\bibitem{Liu}
	Z. Liu, H. Wang, L. Weng, Y. Yang, 
	Ship rotated bounding box space for ship extraction from high-resolution optical satellite images with complex backgrounds, 
	IEEE Geosci. Remote Sens. Lett. 13 (8) (2016) 1074–1078,
	doi: 10.1109/LGRS.2016.2565705.
\bibitem{Bazin}
	J. -C. Bazin, I. Kweon, C. Demonceaux, P. Vasseur, 
	Rectangle extraction in catadioptric images,
	in: IEEE 11th International Conference on Computer Vision, Rio de Janeiro, Brazil, 2007, pp. 1-7, 
	doi: 10.1109/ICCV.2007.4409208.
\bibitem{Stroppa}
	F. Stroppa, C. Loconsole, A. Frisoli,
	Convex polygon fitting in robot-based neurorehabilitation, 
	Appl. Soft Comput. J. 68 (2018) 609-625, 
	doi:10.1016/j.asoc.2018.04.013.
\bibitem{Seo}
	S. Seo, J. Lee, Y. Kim, 
	Extraction of boundaries of rooftop fenced buildings from airborne laser scanning data using rectangle models, IEEE Geosci. Remote Sens. Lett. 11 (2) (2014) 404-408,
	doi: 10.1109/LGRS.2013.2263575.
\bibitem{Feng}
	M. Feng, T. Zhang, S. Li, G. Jin, Y. Xia,
	An improved minimum bounding rectangle algorithm for regularized building boundary extraction from aerial LiDAR point clouds with partial occlusions, 
	Int. J. Remote Sens. 41 (2) (2020) 300-319,
	doi: 10.1080/01431161.2019.1641245.
\bibitem{Yin}
	P. Yin,
	A new method for polygonal approximation using genetic algorithms, 
	Pattern Recognit. Lett. 19 (1998) 1017-1026, 
	doi:10.1016/s0167-8655(98)00082-8.
\bibitem{Yin2}
	P. Yin,
	Genetic algorithms for polygonal approximation of digital curves,
	Int. J. Pattern Recognit. Artif. Intell. 13 (7) (1999) 1061-1082,
	doi:10.1142/S0218001499000598.
\bibitem{Tsai}
	Y. Tsai, 
	Fast polygonal approximation based on genetic algorithms, 
	in: 5th IEEE/ACIS International Conference on Computer and Information Science and 1st IEEE/ACIS International Workshop on Component-Based Software Engineering, Software Architecture and Reuse, 2006, pp. 322–326,
	doi:10.1109/icis-comsar.2006.39.
\bibitem{LiuWatson}
	Z. Liu, J. Watson, A. Allen,
	A polygonal approximation of shape boundaries of marine plankton based-on genetic algorithms,
	J. Vis. Commun. Image Represent. 41 (2016) 305-313,
	doi:10.1016/j.jvcir.2016.10.010.
\bibitem{Liparulo}
	L. Liparulo, A. Proietti, M. Panella, 
	Fuzzy membership functions based on point-to-polygon distance evaluation, 
	in: 2013 IEEE International Conference on Fuzzy Systems, 2013, pp. 1–8,
	doi:10.1109/fuzz-ieee.2013.6622449.
\bibitem{Werman}
	M. Werman, D. Keren, 
	A Bayesian method for fitting parametric and non-parametric models to noisy data, 
	IEEE Trans. Pattern Analysis Mach. Intell. 23 (5) (2001) 528–534,
	doi: 10.1109/34.922710.
\bibitem{Freeman}
	H. Freeman and R. Shapira, 
	Determining the minimum-area encasing rectangle for an arbitrary closed curve, 
	Commun. ACM, 18 (7) (1975) 409–413,
	doi:10.1145/360881.360919.
\bibitem{Chaudhuri2007}
	D. Chaudhuri and A. Samal,
	A simple method for fitting of bounding rectangle to closed regions,
	Pattern Recognit. 40 (7) (2007) 1981-1989,
	doi:10.1016/j.patcog.2006.08.003.
\bibitem{Kwak2012}
	E. Kwak, M. Al-Durgham, A. Habib,
	Automatic 3D building model generation from lidar and image data using sequential minimum bounding rectangle,
	Int. Arch. Photogramm. Remote Sens. Spatial Inf. Sci., XXXIX-B3 (2012) 285–290, doi:10.5194/isprsarchives-XXXIX-B3-285-2012.
\bibitem{Kwak2014}
	E. Kwak, A. Habib,
	Automatic representation and reconstruction of DBM from LiDAR data using Recursive Minimum Bounding Rectangle,
	ISPRS J Photogramm. Remote Sens. 93 (2014) 171-191,
	doi:10.1016/j.isprsjprs.2013.10.003.
\bibitem{Kabolizade}
	M. Kabolizade, H. Ebadi, A. Mohammadzadeh,
	Design and implementation of an algorithm for automatic 3D reconstruction of building models using genetic algorithm,
	Int. J. Appl. Earth Obs. Geoinf. 19 (2012) 104-114,
	doi:10.1016/j.jag.2012.05.006.
\bibitem{WangY}
	Y. Wang, L. Wang, H. Lu, Y. He,
	Segmentation based rotated bounding boxes prediction and image synthesizing for object detection of high resolution aerial images,
	Neurocomputing, 388 (2020) 202-211,
	doi:10.1016/j.neucom.2020.01.039.
\bibitem{Xia}
	G.-S. Xia, X. Bai, J. Ding, Z. Zhu, S. Belongie, J. Luo, M. Datcu, M. Pelillo, L. Zhang, 
	Dota: a large-scale dataset for object detection in aerial images, 
	in: Proceedings of the IEEE Conference on Computer Vision and Pattern Recognition (CVPR), 2018.
\bibitem{Ding}
	J. Ding, N. Xue, Y. Long, G.-S. Xia, Q. Lu,  
	Learning RoI transformer for detecting oriented objects in aerial images,
	in: Proceedings of the IEEE Conference on Computer Vision and Pattern Recognition (CVPR), 2019.
\bibitem{Edgcomb}
	A. Edgcomb, F. Vahid, 
	Automated fall detection on privacy-enhanced video,
	in: 2012 Annual International Conference of the IEEE Engineering in Medicine and Biology Society, San Diego, CA, USA, 2012, pp. 252-255, 
	doi: 10.1109/EMBC.2012.6345917.
\bibitem{Hu}
	W.-C. Hu, C.-H. Chen, T.-Y. Chen, D.-Y. Huang, Z.-C. Wu,
	Moving object detection and tracking from video captured by moving camera,
	J. Vis. Commun. Image Represent. 30 (2015) 164-180,
	doi:10.1016/j.jvcir.2015.03.003.
\bibitem{Chaudhuri2011}
	D. Chaudhuri, N. K. Kushwaha, I. Sharif, A. Samal,
	Finding best-fitted rectangle for regions using a bisection method, 
	Mach. Vis. Appl. 23 (6) (2011) 1263–1271, 
	doi:10.1007/s00138-011-0348-6.
\bibitem{Sampath}
	A. Sampath, J. Shan, 
	Building boundary tracing and regularization from airborne lidar point clouds,
	Photogramm. Eng. Remote Sens. 73 (7) (2007) 805–812,
	doi:10.14358/PERS.73.7.805.
\bibitem{Sinnreich}
	J. Sinnreich,
	Least-squares fitting of polygons, 
	Pattern Recognit. Image Analysis 26 (2) (2016) 343-349, 
	doi:10.1134/S10546618160202180.



\end{thebibliography}

\section*{Acknowledgement}
This work was supported by 2021 Science and Technology Projects of Zhejiang Provincial Department of Natural, which is a joint project with Ningbo Tianyi Design Research of Surveying and Mapping Co. Ltd.

\end{document}